\documentclass[10pt, a4paper]{article}
\usepackage{lrec}
\usepackage{graphicx}
\usepackage{tabularx}
\usepackage{soul}
\usepackage{epstopdf}

\usepackage{hyperref}
\usepackage{xstring}
\usepackage{amsmath}

\usepackage{graphicx}
\usepackage{caption}
\usepackage{subcaption}
\usepackage{multirow}

\title{``A Passage to India'': Pre-trained Word Embeddings for Indian Languages}
\name{Kumar Saurav\textsuperscript{$\dagger$}, Kumar Saunack\textsuperscript{$\dagger$}, Diptesh Kanojia\textsuperscript{$\dagger$,$\clubsuit$,$\star$}, and Pushpak Bhattacharyya\textsuperscript{$\dagger$}}

\address{\textsuperscript{$\dagger$}Indian Institute of Technology Bombay, India\\
\textsuperscript{$\clubsuit$}IITB-Monash Research Academy, India\\
\textsuperscript{$\star$}Monash University, Australia\\
         \{krsrv, krsaunack, diptesh, pb\}@cse.iitb.ac.in\\}

\abstract{
Dense word vectors or `word embeddings' which encode semantic properties of words, have now become integral to NLP tasks like Machine Translation (MT), Question Answering (QA), Word Sense Disambiguation (WSD), and Information Retrieval (IR). In this paper, we use various existing approaches to create multiple word embeddings for 14 Indian languages. We place these embeddings for all these languages, \textit{viz.}, Assamese, Bengali, Gujarati, Hindi, Kannada, Konkani, Malayalam, Marathi, Nepali, Odiya, Punjabi, Sanskrit, Tamil, and Telugu in a single repository. Relatively newer approaches that emphasize catering to context (BERT, ELMo, \textit{etc.}) have shown significant improvements, but require a large amount of resources to generate usable models. We release pre-trained embeddings generated using both contextual and non-contextual approaches. We also use MUSE and XLM to train cross-lingual embeddings for all pairs of the aforementioned languages. To show the efficacy of our embeddings, we evaluate our embedding models on XPOS, UPOS and NER tasks for all these languages. We release a total of 436 models using 8 different approaches. We hope they are useful for the resource-constrained Indian language NLP. The title of this paper refers to the famous novel ``A Passage to India'' by E.M. Forster, published initially in 1924.   \\ \newline  }

\begin{document}

\maketitleabstract

\section{Introduction}

India has a total of 22 scheduled languages with a combined total of more than a billion speakers. Indian language content on the web is accessed by approximately 234 million speakers across the world\footnote{\href{https://home.kpmg/in/en/home/insights/2017/04/indian-language-internet-users.html}{Source Link}}. Despite the enormous user base, Indian languages are known to be low-resource or resource-constrained languages for  NLP. Word embeddings have proven to be important resources, as they provide a dense set of features for downstream NLP tasks like MT, QA, IR, WSD, \textit{etc.} Unlike in classical Machine Learning wherein features have at times to be extracted in a supervised manner, embeddings can be obtained in a completely unsupervised fashion. For Indian languages, there are little corpora and few datasets of appreciable size available for computational tasks. The wikimedia dumps which are used for generating pre-trained models are insufficient. Without sufficient data, it becomes difficult to train embeddings.

NLP tasks that benefit from these pre-trained embeddings are very diverse. Tasks ranging from word analogy and spelling correction to more complex ones like Question Answering \cite{bordes2014question}, Machine Translation \cite{artetxe2019effective},  and Information Retrieval \cite{diaz2016query} have reported improvements with the use of well-trained embeddings models. The recent trend of transformer architecture based neural networks has inspired various language models that help train contextualized embeddings \cite{devlin2018bert,peters2018deep,melamud2016context2vec,lample2019cross}. They report significant improvements over various NLP tasks and release pre-trained embeddings models for many languages. One of the shortcomings of the currently available pre-trained models is the corpora size used for their training. Almost all of these models use Wikimedia corpus to train models which is insufficient for Indian languages as Wikipedia itself lacks significant number of articles or text in these languages. Although there is no cap or minimum number of documents/lines which define a usable size of a corpus for training such models, it is generally considered that the more input training data, the better the embedding models.

Acquiring raw corpora to be used as input training data has been a perennial problem for NLP researchers who work with low resource languages. Given a raw corpus, monolingual word embeddings can be trained for a given language. Additionally, NLP tasks that rely on utilizing common linguistic properties of more than one language need cross-lingual word embeddings, \textit{i.e.,} embeddings for multiple languages projected into a common vector space. These cross-lingual word embeddings have shown to help the task of cross-lingual information extraction \cite{levy2017zero}, False Friends and Cognate detection \cite{merlo2019cross}, and Unsupervised Neural Machine Translation \cite{artetxe2018unsupervised}. With the recent advent of contextualized embeddings, a significant increase has been observed in the types of word embedding models. It would be convenient if a single repository existed for all such embedding models, especially for low-resource languages. Our work creates such a repository for fourteen Indian languages, keeping this in mind, by training and deploying 436 models with different training algorithms (like word2vec, BERT, etc.) and hyperparameters as detailed further in the paper. \textit{Our key contributions are: \\(1) We acquire raw monolingual corpora for fourteen languages, including Wikimedia dumps. (2) We train various embedding models and evaluate them. (3) We release these embedding models and evaluation data in a single repository\footnote{\href{http://www.cfilt.iitb.ac.in/~diptesh/embeddings}{Repository Link}}.} 

The roadmap of the paper is as follows: in section 2, we discuss previous work; section 3 discusses the corpora and our evaluation datasets; section 4 briefs on the approaches used for training our models, section 5 discusses the resultant models and their evaluation; section 6 concludes the paper.

\section{Literature Survey}
Word embeddings were first introduced in \cite{bengio2003} when it was realised that learning the joint probability of sequences was not feasible due to the \textit{`curse of dimensionality'}, \textit{i.e.,} at that time, the value added by an additional dimension seemed much smaller than the overhead it added in terms of computational time, and space.
Since then, several developments have occurred in this field. Word2Vec \cite{mikolov2013efficient} showed the way to train word vectors. The models introduced by them established new state-of-the-art on tasks such as Word Sense Disambiguation (WSD). GloVE \cite{pennington2014glove} and FastText \cite{bojanowski2017enriching} further improved on results shown by \newcite{mikolov2013efficient}, where GloVE used a co-occurrence matrix and FastText utilized the sub-word information to generate word vectors. Sent2Vec \cite{pagliardini2017unsupervised} generates sentence vectors inspired by the same idea. Universal Sentence Embeddings \cite{cer2018universal}, on the other hand, creates sentence vectors using two variants: transformers and DANs. Doc2Vec \cite{le2014distributed} computes a feature vector for every document in the corpus. Similarly, Context2vec \cite{melamud2016context2vec} learns embedding for variable length sentential context for target words.

The drawback of earlier models was that the representation for each word was fixed regardless of the context in which it appeared. To alleviate this problem, contextual word embedding models were created. ELMo \cite{peters2018deep} used bidirectional LSTMs to improve on the previous works. Later, BERT\cite{devlin2018bert} used the transformer architecture to establish new a state-of-the-art across different tasks. It was able to learn deep bidirectional context instead of just two unidirectional contexts, which helped it outperform previous models. XLNet \cite{yang2019xlnet} was a further improvement over BERT. It addressed the issues in BERT by introducing permutation language modelling, which allowed it to surpass BERT on several tasks.

Cross-lingual word embeddings, in contrast with monolingual word embeddings, learn a common projection between two monolingual vector spaces. MUSE \cite{conneau2017word} was introduced to get cross-lingual embeddings across different languages. VecMap\cite{artetxe2018acl} introduced unsupervised learning for these embeddings. BERT, which is generally used for monolingual embeddings, can also be trained in a multilingual fashion. XLM\cite{lample2019cross} was introduced as an improvement over BERT in the cross-lingual setting.

The official repository for FastText has several pretrained word embedding for multiple languages, including some Indian languages. The French, Hindi and Polish word embeddings, in particular, have been evaluated on Word Analogy datasets, which were released along with the paper. \newcite{haider-2018-urdu} release word embeddings for the Urdu language, which is one of the Indian languages we do not cover with this work. To evaluate the quality of embeddings, they were tested on Urdu translations of English similarity datasets.

\section{Dataset and Experiment Setup}
We collect pre-training data for over 14 Indian languages (from a total of 22 scheduled languages in India), including Assamese (as), Bengali (bn), Gujarati (gu), Hindi (hi), Kannada (kn), Konkani (ko), Malayalam (ml), Marathi (mr), Nepali (ne), Odiya (or), Punjabi (pa), Sanskrit (sa), Tamil (ta) and Telugu (te). These languages account for more than 95\% of the entire Indian population, with the most widely spoken language, Hindi, alone contributing 43\% to the figure\footnote{\url{https://en.wikipedia.org/wiki/List_of_languages_by_number_of_native_speakers_in_India}}. Nonetheless, data that is readily available for computational purposes has been excruciatingly limited, even for these 14 languages.

One of the major contributions of this paper is the accumulation of data in a single repository. This dataset has been collected from various sources, including ILCI corpora \cite{choudhary2011creating,bansal2013corpora}, which contains parallel aligned corpora (including English) with Hindi as the source language in tourism and health domains. As a baseline dataset, we first extract text from Wikipedia dumps\footnote{As on 15th August, 2019}, and then append the data from other sources onto it. We added the aforementioned ILCI corpus, and then for Hindi, we add the monolingual corpus from HinMonoCorp 0.5 \cite{bojar2014hindencorp}, increasing the corpus size by 44 million sentences. For Hindi, Marathi, Nepali, Bengali, Tamil, and Gujarati, we add crawled corpus of film reviews and news websites\footnote{\url{https://github.com/goru001}}. For Sanskrit, we download a raw corpus of prose\footnote{\url{http://sanskrit.jnu.ac.in/currentSanskritProse/}} and add it to our corpus. Further, we describe the preprocessing and tokenization of our data.

\subsection{Preprocessing}
The corpora collected is intended to be set in general domain instead of being domain-specific, and hence we start by collecting general domain corpora via Wikimedia dumps. We also add corpora from various crawl sources to respective individual language corpus.
All the corpora is then cleaned, with the first step being the removal of HTML tags and links which can occur due to the presence of crawled data. Then, foreign language sentences (including English) are removed from each corpus, so that the final pre-training corpus contains words from only its language. Along with foreign languages, numerals written in any language are also removed. Once these steps are completed, paragraphs in the corpus are split into sentences using sentence end markers such as full stop and question mark. Following this, we also remove any special characters which may have included punctuation marks (example - hyphens,  commas etc.).

The statistics for the resulting corpus are listed in Table \ref{tab:corpus}.

\begin{table}[!ht]
    \centering
    \begin{tabular}{|c|c|c|c|}
        \hline
         \textbf{Language} & \textbf{Abbr.} & \textbf{Sentences} & \textbf{Words} \\
         \hline
         Hindi & hin & 48,115,256 & 3,419,909 \\
         Bengali & ben & 1,563,137 & 707,473 \\
         Telugu & tel & 1,019,430 & 1,255,086\\
         Tamil & tam & 881,429 & 1,407,646\\
         Nepali & nep & 705,503 & 314,408\\
         Sanskrit & san & 553,103 & 448,784\\
         Marathi & mar & 519,506 & 498,475\\
         Punjabi & pan & 503,330 & 247,835\\
         Malayalam & mal & 493,234 & 1,325,212\\
         Gujarati & guj & 468,024 & 182,566\\
         Konkani & knn & 246,722 & 76,899\\
         Oriya & ori & 112,472 & 55,312\\
         Kannada & kan & 51,949 & 30,031\\
         Assamese & asm & 50,470 & 29,827\\
         \hline
    \end{tabular}
    \caption{Corpus statistics for each Indic language with their ISO 639-3 abbreviations (total number of sentences and words)}
    \label{tab:corpus}
\end{table}
% Flair for non-contextual embeddings
% perplexity scores for elmo models
% NER scores for Indian languages
% word analogy dataset creation
\begin{figure*}[h]
\centering
\begin{subfigure}{0.5\textwidth}
  \centering
  \includegraphics[width=\linewidth]{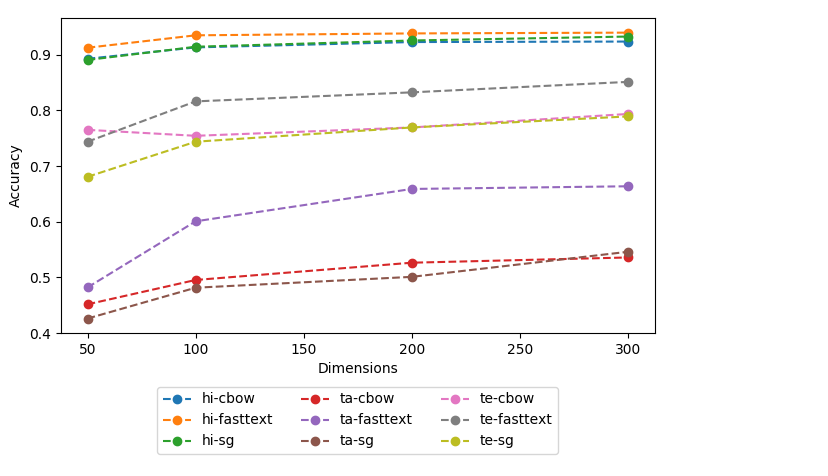}
  \caption{Performance on UPOS tagged dataset}
  \label{img:upos-perf}
\end{subfigure}%
\begin{subfigure}{0.5\textwidth}
  \centering
  \includegraphics[width=\linewidth]{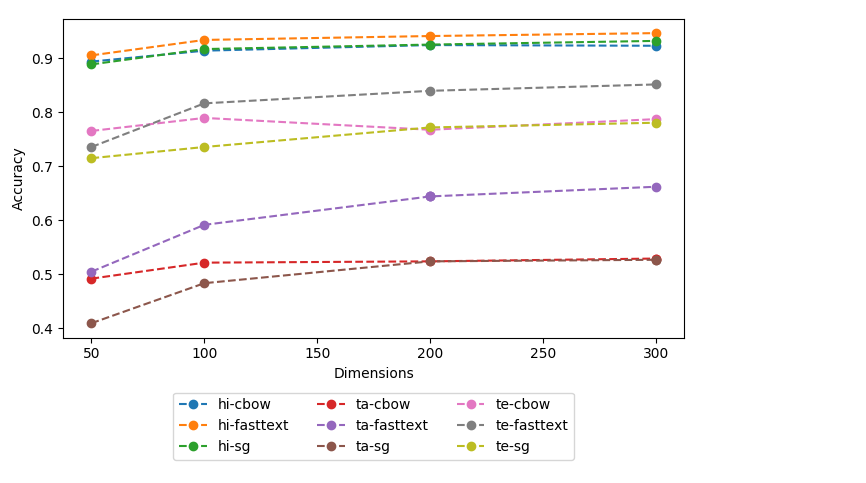}
  \caption{Performance on XPOS tagged dataset}
  \label{img:xpos-perf}
\end{subfigure}
\caption{Performance of skip-gram, CBOW, and fasttext models on POS tagging task. Plotted graph is Accuracy vs Dimension. Legend is "language"-"model". Note that FastText is the best performer in each case, and learning saturates around 200 dimensions}
\vspace{0.5cm}
% \label{fig:test}
\end{figure*}

\begin{figure}[h]
    \centering
      \includegraphics[width=\linewidth]{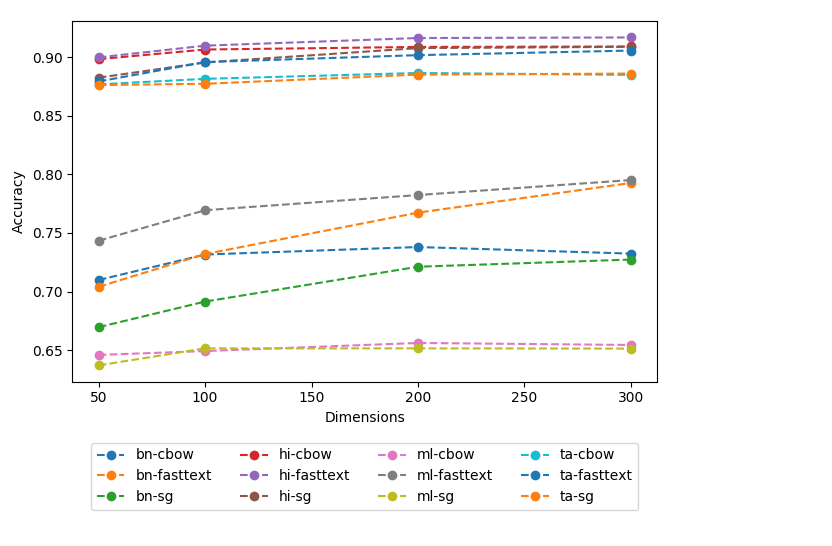}
      \caption{Performance of skip-gram,  CBOW, and fasttext models on NER tagged dataset}
      \label{img:ner-perf}
\end{figure}

\subsection{Experiment Setup}
There is a prevailing scarcity of standardised benchmarks for testing the efficacy of various word embedding models for resource-poor languages. We conducted experiments across some rare standardised datasets that we could find and created new evaluation tasks as well to test the quality of non-contextual word embeddings. The Named Entity Recognition task, collected from \cite{murthy2018improving}, and FIRE 2014 workshop for NER, contains NER tagged data for 5 Indian languages, namely Hindi, Tamil, Bengali, Malayalam, and Marathi. We also use a Universal POS (UPOS), as well as an XPOS (language-specific PoS tags) tagged dataset, available from the  Universal Dependency (UD) treebank \cite{nivre2016universal}, which contains POS tagged data for 4 Indian languages, Hindi, Tamil, Telugu, and Marathi. 

For the tasks of NER, UPOS tagging, XPOS tagging, we use the Flair library \cite{akbik2018coling}, which embeds our pre-trained embeddings as inputs for training the corresponding tagging models. The tagging models provided by Flair are vanilla BiLSTM-CRF sequence labellers. For the task of word analogy dataset, we simply use the vector addition and subtraction operators to check accuracy (\textit{i.e.,} $v(\text{France}) - v(\text{Paris}) + v(\text{Berlin})$ should be close to $v(\text{Germany})$).

For contextual word embeddings, we collect the statistics provided at the end of the pre-training phase to gauge the quality of the embeddings - perplexity scores for ELMo, masked language model accuracy for BERT, and so on. We report these values in Table \ref{tab:elmo}.

% We also create a new evaluation metric for word analogy using IndoWordnet \cite{bhattacharyya2017indowordnet} to extract hypernyms (more specifically IS\_A relation) for the most common words in the corpus. Each pair in the set of all such relations constitute an entry in our word analogy dataset. We are making this dataset public as one of the contributions of this paper. We have, however, been unable to test the embeddings on this metric though.

\section{Models and Evaluation}

\begin{table*}[!ht]
    \centering
    \begin{tabular}{|c|c|c|c|c|c|c|c|c|c|c|c|c|c|c|}
    \hline
    \textbf{Language} & as & bn & gu & hi & ml & mr & kn & ko & ne & or & pa & sa & ta & te \\ \hline
    \textbf{Perplexity} & 455 & 354 & 183 & 518 & 1689 & 522 & 155368 & 325 & 253 & 975 & 145 & 399 & 781 & 82 \\ \hline
    \end{tabular}
    \caption{ELMo prerplexity scores}
    \vspace{0.6cm}
    \label{tab:elmo}
\end{table*}

In this section, we briefly describe the models created using the approaches mentioned above in the paper.

\subsection{Word2Vec (skip-gram and CBOW)}
Word2Vec embeddings \cite{mikolov2013distributed} of dimensions \{50, 100, 200, 300\} for both skip-gram and CBOW architectures are created using the gensim library \cite{rehurek_lrec} implementation of Word2Vec. Words with a frequency less than 2 in the entire corpus are treated as unknown (out-of-vocabulary) words. For other parameters, default settings of gensim are used. There are no pre-trained Word2Vec word embeddings for any of the 14 languages available publicly.

\subsection{FastText}
FastText embeddings \cite{bojanowski2017enriching} of dimensions \{50, 100, 200, 300\} (skip-gram architecture) were created using the gensim library \cite{rehurek_lrec} implementation of FastText. Words with a frequency less than 2 in the entire corpus are treated as unknown (out-of-vocabulary) words. For other parameters, default settings of gensim are used. Except for Konkani and Punjabi, the official repository for FastText provides pre-trained word embeddings for the Indian languages. However, we have trained our word embeddings on a much larger corpus than those used by FastText.

\subsection{GloVe}
We create GloVe embeddings \cite{pennington2014glove} of dimensions \{50, 100, 200, and 300\}. Words with occurrence frequency less than 2 are not included in the library. The co-occurrence matrix is created using a symmetric window of size 15.  There are no pre-trained word embeddings for any of the 14 languages available with the GloVE embeddings repository\footnote{\url{https://nlp.stanford.edu/projects/glove/}}. We create these models and provide them with our repository.

\subsection{MUSE}
MUSE embeddings are cross-lingual embeddings that can be trained using the fastText embeddings, which we had created previously. Due to resource constraints and the fact that cross-lingual representations require a large amount of data, we choose to train 50-dimensional embeddings for each language pair. We train for all the language pairs (14*14) and thus produce 196 models using this approach and provide them in our repository. The training for these models took 2 days over 1 x 2080Ti GPU (12 GB).

\subsection{ELMo}
We train ELMo embeddings \cite{peters2018deep} of 512 dimensions. These vectors are learned functions of the internal states of a deep bidirectional language model (biLM). The training time for each language corpus was approximately 1 day on a 12 GB Nvidia GeForce GTX TitanX GPU. The batch size is reduced to 64, and the embedding model was trained on a single GPU. The number of training tokens was set to tokens multiplied by 5. We choose this parameter based on the assumption that each sentence contains an average of 4 tokens. There are no pre-trained word embeddings for any of the 14 languages available on the official repository. We provide these models in our repository.

\subsection{BERT}
We train BERT (Bidirectional Encoder Representations from Transformers) embeddings \cite{devlin2018bert} of 300 dimensions. Since BERT can be used to train a single multilingual model, we combine and shuffle corpora of all languages into a single corpus and used this as the pre-training data. We use sentence piece embeddings \cite{SentencePiece} that we trained on the corpus with a vocabulary size of 25000. Pre-training this model was completed in less than 1 day using 3 * 12 GB Tesla K80 GPUs. The official repository for BERT provides a multilingual model of 102 languages, which includes all but 4 (Oriya, Assamese, Sanskrit, Konkani) of the 14 languages. We provide a single multilingual BERT model for all the 14 languages, including these 4 languages.

\subsection{XLM}

We train cross-lingual contextual BERT representation language model using the XLM git repository\footnote{\url{https://github.com/facebookresearch/XLM}}. We train this model for 300-dimensional embeddings and over the standard hyperparameters as described with their work. The corpus vocabulary size of 25000 was chosen. We use a combined corpus of all 14 Indian languages and shuffle the sentences for data preparation of this model. We use the monolingual model (MLM) method to prepare data as described on their Git repository. This model also required the Byte-pair encoding representations as input and we train them using the standard fastBPE implementation as recommended over their Github. The training for this model took 6 days and 23 hours over 3 x V100 GPUs (16 GB each). 

\subsection{Evaluation} 
\label{eval}
We evaluate and compare the performance of FastText, Word2Vec, and GloVE embedding models on UPOS and XPOS datasets. The results are shown in the image  \ref{img:upos-perf} and in \ref{img:xpos-perf}, respectively. The performance of non-contextual word embedding models on NER dataset is shown in image \ref{img:ner-perf}. The perplexity scores for ELMo training are listed in table \ref{tab:elmo}. We observe that FastText outperforms both GloVE and Word2Vec models. For Indian languages, the performance of FastText is also an indication of the fact that morphologically rich languages require embedding models with sub-word enriched information. This is clearly depicted in our evaluation.

The overall size of all the aforementioned models was very large to be hosted on a Git repository. We host all of these embeddings in a downloadable ZIP format each on our server, which can be accessed via the link provided above.

\section{Results and Discussion}
We have created a comprehensive set of standard word embeddings for multiple Indian languages. We release a total of 422 embedding models for 14 Indic languages. The models contain 4 varying dimensions (50, 100, 200, and 300) each of GloVE, Skipgram, CBOW, and FastText; 1 each of ELMo for every language; a single model each of BERT and XLM of all languages. They also consist of 182 cross-lingual word embedding models for each pair. However, due to the differences in language properties as well as corpora sizes, the quality of the models vary. Table \ref{tab:corpus} shows the language wise corpus statistics. Evaluation of the models has already been presented in Section \ref{eval}. An interesting point is that even though Tamil and Telugu have comparable corpora sizes, the evaluations of their word embeddings show different results. Telugu models consistently outperform Tamil models on all common tasks. Note that the NER tagged dataset was not available for Telugu, so they could not be compared on this task.

This also serves to highlight the difference between the properties of these two languages. Even though they belong to the same language family, \textit{Dravidian}, and their dataset size is the same, their evaluations show a marked difference. Each language has 3 non-contextual embeddings (word2vec-skipgram, word2vec-cbow and fasttext-skipgram), and a contextual embedding (ElMo). Along with this, we have created multilingual embeddings via BERT. For BERT pre-training, the masked language model accuracy is 31.8\% and next sentence prediction accuracy is 67.9\%. Cross-lingual embeddings, on the other hand, have been created using XLM and MUSE.

\section{Conclusion and Future Work}

The recent past has seen tremendous growth in NLP with ElMo, BERT and XLNet being released in quick succession. All such advances have improved the state-of-the-art in various tasks like NER, Question Answering, Machine Translation, etc. However, most of these results have been presented predominantly for a single language- English. With the potential that Indian languages computing has, it becomes pertinent to perform research in word embeddings for local, low-resource languages as well. In this paper, we present the work done on creating a single repository of corpora for 14 Indian languages. We also discuss the creation of different embedding models in detail. As for our primary contribution, these word embedding models are being publicly released.\\

In the future, we aim to refine these embeddings and do a more exhaustive evaluation over various tasks such as POS tagging for all these languages, NER for all Indian languages, including a word analogy task. Presently evaluations have been carried out on only a few of these tasks. Also, with newer embedding techniques being released in quick successions, we hope to include them in our repository. The model's parameters can be trained further for specific tasks or improving their performance in general. We hope that our work serves as a stepping stone to better embeddings for low-resource Indian languages. 

\section{Bibliographical References}
\label{main:ref}
\bibliographystyle{lrec}
\bibliography{lrec2020W-xample}

\end{document}